%% file: acl_latex.tex
\documentclass[11pt]{article}

\usepackage[preprint]{acl}

\usepackage{times}
\usepackage{latexsym}
\usepackage[T1]{fontenc}
\usepackage[utf8]{inputenc}
\usepackage{microtype}
\usepackage{inconsolata}
\usepackage{graphicx}

\usepackage{amsmath}
\usepackage{amssymb}
\usepackage{amsfonts}
\usepackage{bm}
\usepackage{booktabs}
\usepackage{multirow}
\usepackage{array}
\usepackage{xcolor}
\usepackage{enumitem}
\usepackage{xspace}
\usepackage{comment}
\usepackage[table]{xcolor}

\definecolor{bestbg}{RGB}{250,238,238}
\definecolor{secondbg}{RGB}{238,244,250}
\definecolor{thirdbg}{RGB}{239,247,239}

\newcommand{\best}[1]{\cellcolor{bestbg}\textbf{#1}}
\newcommand{\second}[1]{\cellcolor{secondbg}\underline{#1}}
\newcommand{\third}[1]{\cellcolor{thirdbg}\textit{#1}}

\usepackage{amsthm}

\usepackage[dvipsnames]{xcolor}

 \usepackage{textcomp}
 \usepackage[most]{tcolorbox}   

 \newtcblisting[auto counter]{promptbox}[2]{%
   enhanced, breakable, listing only,
  colback=black!2, colframe=black!45, boxrule=0.4pt, arc=1.2pt,
   left=4pt, right=4pt, top=3pt, bottom=3pt,
   title={Prompt~\thetcbcounter: #1}, label={#2},
   fonttitle=\small\bfseries, coltitle=black, colbacktitle=black!10,
   listing options={basicstyle=\ttfamily\footnotesize, breaklines=true,
     breakatwhitespace=true, columns=fullflexible, keepspaces=true,
     showstringspaces=false, upquote=true}}

\newcommand{\method}{\textsc{Pro-Memory}\xspace}

\title{Prospective Memory for Proactive LLM Agents}

\author{Yuanchen Bei$^1$,\ Zhining Liu$^1$,\ Tianxin Wei$^1$,\ Weizhi Zhang$^2$,\ Zhichen Zeng$^1$,\\ {\bf Philip S. Yu}$^2${\bf,\ Hendrik Hamann}$^{3,4}${\bf,\ Jingrui He}$^1${\bf,\ Hanghang Tong}$^1$ \\
$^1$University of Illinois Urbana-Champaign\quad $^2$University of Illinois Chicago\\
$^3$Stony Brook University\quad $^4$Brookhaven National Laboratory}

\begin{document}
\maketitle

\begin{abstract}
\input{contents/abstract}
\end{abstract}

\section{Introduction}
\label{sec:intro}
\input{contents/intro}

\section{Related Work}
\label{sec:related}
\input{contents/related}

\section{\method}
\label{sec:method}
\input{contents/method}

\section{Experiments}
\label{sec:exp}
\input{contents/exper}

\section{Conclusion}
\input{contents/conclusion}

\section*{Limitations}
\label{sec:limits}
Our current study primarily focuses on text-based proactive interactions, while real-world agents may need to recognize future opportunities from multimodal signals, such as images, audio, and video.
Extending prospective memory to multimodal intention formation and cross-modal cue recognition is a promising direction.
Furthermore, exploring more complex long-horizon scenarios involving interdependent intentions and evolving user needs remains an important avenue for future work.

\section*{Ethics Statement}
This work studies proactive assistance using previously released benchmarks for evaluation, without deploying autonomous actions to real users. Proactive agents may nevertheless misinterpret user intentions, interrupt ongoing activities, or expose sensitive contextual information through unwanted interventions. Our approach aims to ground interventions in available evidence, but benchmark improvements do not establish safety in real-world deployments. Practical systems should provide user control over proactive behavior and incorporate appropriate privacy safeguards and human oversight.


\bibliography{custom}
\clearpage
\appendix

\input{contents/appendix}

\end{document}

%% file: contents/abstract.tex
Most memory systems for LLM agents are \emph{retrospective}, preserving and retrieving past information to inform current decisions.
Proactive agents, however, also require \emph{\textbf{prospective memory}}: retaining future-directed intentions and realizing them when appropriate opportunities arise, even without explicit requests.
Crucially, \emph{remembering the right intention does not guarantee acting at the right time}: a related observation may call for waiting, while a later cue may make the same intention actionable.
Inspired by cognitive theories of human prospective memory, we introduce \textbf{\method}, a memory framework that separates maintaining future intentions from deciding when and how to realize them.
\method preserves pending objectives linked to their supporting evidence, assesses new observations against their original goals and activation conditions, and coordinates appropriate contributions with the agent's ongoing actions.
It further updates intention states based on actions actually issued or evidence of resolution, while an experience memory distills feedback from completed episodes into guidance for future decisions.
Experiments on two benchmarks spanning delayed-intention tasks and open-ended proactive assistance demonstrate consistent improvements in proactive task performance over representative memory approaches across multiple LLM backbones, while limiting unnecessary interventions.

%% file: contents/intro.tex
Large language model (LLM) agents increasingly rely on memory to preserve user-specific information and interaction experience beyond a single context window~\citep{zhang2025survey,huang2026survey}.
Existing memory systems summarize past interactions, construct user profiles, or organize experience into structured stores~\citep{zhong2024memorybank,fang2026lightmem,xu2026mem}.
Despite their architectural differences, these approaches primarily address a \emph{retrospective memory} problem: \textit{Given the current context, what past information should be retrieved to inform the present decision?}

\begin{figure}[t]
    \centering
    \includegraphics[width=\linewidth]{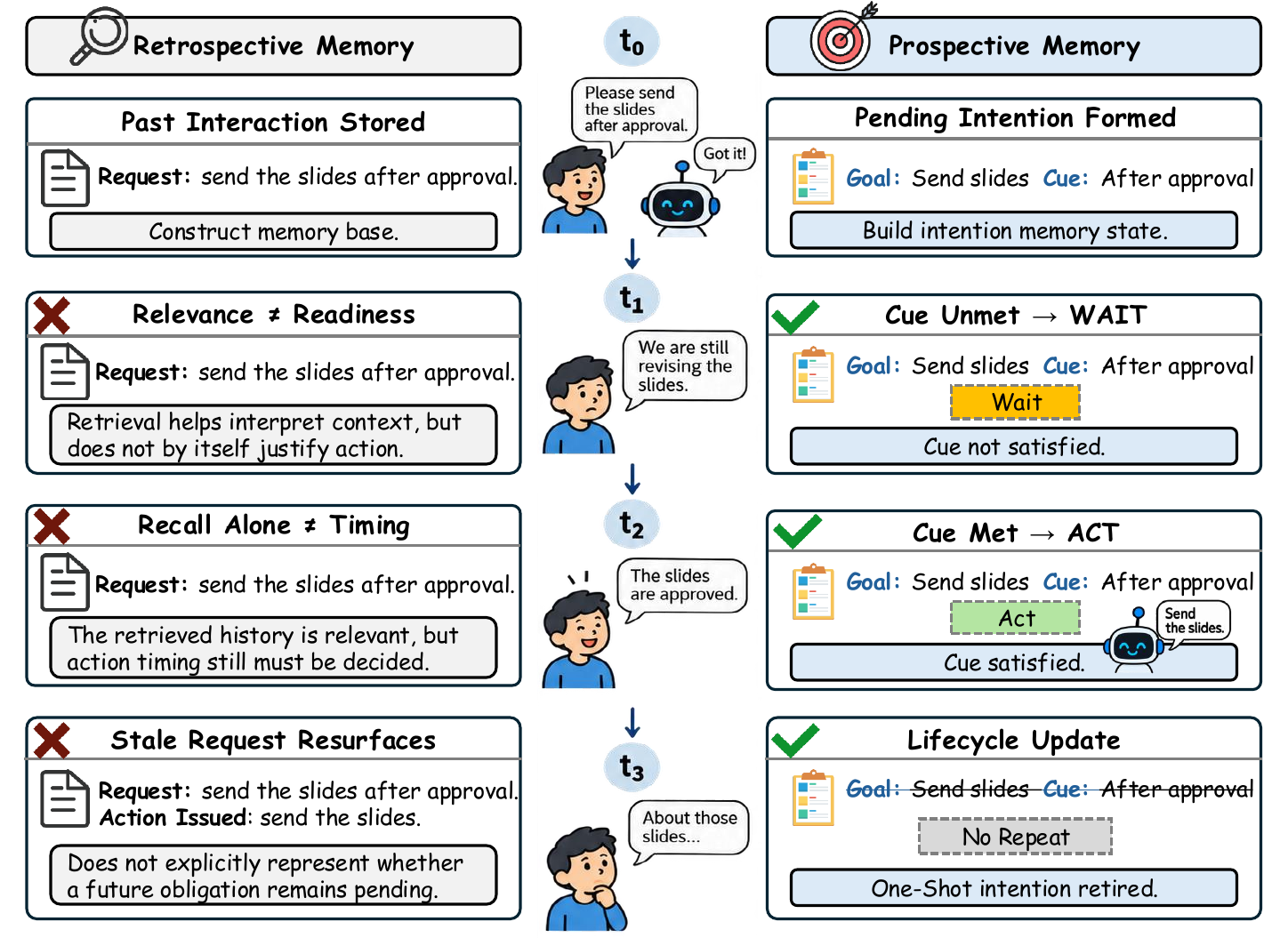}
    \caption{From retrospective recall to prospective realization. While retrospective memory retrieves relevant information, retrieval alone does not determine when to act. Prospective memory instead maintains pending intentions, recognizes actionable opportunities, and updates their lifecycle to support suitable interventions.}
    \label{fig:intro}
\end{figure}

Proactive LLM agents face a different challenge.
They continuously observe evolving contexts and may need to provide assistance before an explicit user request~\citep{lu2025proactive,yang2026contextagent}.
An intention formed earlier may remain unresolved across multiple interactions, becoming actionable only after a future event or change in circumstances.
Acting prematurely can be disruptive, while overlooking an appropriate opportunity can leave the user's needs unmet.
Agents must therefore remember \emph{to act}, not merely \emph{what happened}.
This calls for \emph{prospective memory}: retaining future-directed intentions and realizing them when appropriate opportunities arise~\citep{rummel2023current,liu2026pm}.
For proactive agents, the central question becomes: \textit{Which remembered intention, if any, warrants action now?}

Crucially, \emph{remembering the right intention does not guarantee acting at the right time}.
Consider a user asking an agent to share slides after approval, as illustrated in Figure~\ref{fig:intro}.
A later discussion about revisions may bring the request to mind, but sharing the slides would still be premature.
A subsequent approval notice, however, may make the same intention actionable without repeating the original request.
Once a one-shot contribution has been issued, the corresponding intention should no longer trigger that contribution again.
The challenge is therefore not simply retrieving relevant history, but determining whether a future intention remains pending and whether the current situation warrants action~\citep{zhang2026triggerbench}.
Recent approaches move beyond passive retrieval by anticipating future memory needs~\citep{guo2026t} or selectively providing memory-grounded reminders~\citep{wu2026remember}.
Nevertheless, \emph{surfacing the right memory is not equivalent to realizing the right intention}.
The latter requires preserving \emph{what remains intended}, recognizing \emph{when to act}, and deciding \emph{how to intervene appropriately}.

Cognitive research on human prospective memory provides a natural foundation for addressing this challenge.
\citet{kliegel2002complex} characterize prospective remembering through four processes: intention formation, retention, initiation, and execution.
Importantly, recognizing a relevant cue and retrieving an intention are distinct from coordinating its execution with ongoing activity~\citep{marsh2002dynamics}.
Furthermore, previously completed intentions may still be reactivated and induce erroneous repeated responses~\citep{scullin2012whoops}.
Together, these findings suggest that successful prospective remembering involves more than retaining future intentions: it also requires recognizing appropriate opportunities and regulating how intentions influence ongoing behavior.
For proactive agents, this challenge is further complicated by the fact that the appropriate action may not be known when an intention is first formed, but instead depends on the evidence and capabilities available later.

Guided by these cognitive processes, we introduce \method, a prospective memory framework that separates \emph{maintaining future intentions} from \emph{realizing them in context}. 
Following the four phases of prospective remembering, \underline{\textbf{Formation}} identifies future objectives grounded in user needs, while \underline{\textbf{Maintenance}} preserves unresolved intentions and their supporting evidence across interactions.
\underline{\textbf{Cue-Guided Initiation}} examines historical intention evidence alongside current observations to decide whether to act, seek additional information, or wait, and determines an appropriate contribution when action is warranted.
\underline{\textbf{Selective Enactment}} then considers these memory-driven contributions together with the agent's own action proposals, determines which interventions to actually issue, and updates the intention lifecycle accordingly.
A feedback-driven experience memory provides reusable decision guidance from completed episodes to improve subsequent opportunity assessments.
By determining concrete contributions when appropriate opportunities arise, \method supports both delayed-intention tasks and open-ended proactive assistance.
Experiments on two complementary benchmarks across representative LLM backbones demonstrate \method improves proactive task performance over existing memory approaches while limiting unnecessary interventions.

Our contributions are summarized as follows:

\begin{itemize}[leftmargin=*]

\item \textbf{Problem Formulation.}
We formulate prospective memory for proactive LLM agents as maintaining pending objectives and deciding when and how to act on them in an evolving context.

\item \textbf{New Method.}
We propose \method, a cognitively inspired framework that integrates evidence-linked intention maintenance, cue-guided opportunity assessment, and selective enactment coordinated with ongoing agent actions, complemented by feedback-driven experience memory.

\item \textbf{Comprehensive Evaluation.}
We conduct extensive experiments on two benchmarks spanning delayed-intention tasks and open-ended proactive assistance, demonstrating the effectiveness of \method across LLM backbones.

\end{itemize}

%% file: contents/related.tex
\subsection{Proactive LLM Agents}
Proactive LLM agents aim to anticipate and satisfy user needs before an explicit request, requiring them to continuously perceive evolving contexts and decide both \emph{whether} and \emph{when} to intervene~\citep{yang2026contextagent,lu2025proactive,tang2026proagentbench}. Unlike conventional request-response agents, proactive agents must infer latent user needs from ongoing interactions and act only when sufficient evidence indicates that assistance is timely. This makes proactive assistance inherently temporally dependent: a need may emerge gradually, remain unresolved across multiple interactions, or become actionable only after a later change in user state or context~\cite{kong2026proactivemobile,liu2026proactive}. As a result, memory becomes an important component for preserving relevant information across time and supporting proactive decisions over extended user-agent interactions~\cite{xie2026pask}.

\subsection{Memory for LLM Agents}
Memory has become a central component of LLM agents, enabling them to retain information beyond the immediate context window and reuse past interactions and experience with structured memory management~\citep{xu2026mem,zhong2024memorybank,yang2026plugmem,fang2026lightmem}. Most existing systems are retrospective: they store and organize past information and later retrieve relevant content for a current decision given a deterministic query~\cite{hu2025memory,salama2025meminsight}. 
Recent approaches move beyond passive retrieval by anticipating future memory needs~\citep{guo2026t} or selectively injecting memory-grounded reminders into ongoing execution~\citep{wu2026remember}.
More recently, PIS~\citep{zhao2026making} organizes deferred intentions as typed trigger-action records and applies structural rules to filter eligible intentions.
In contrast, \method maintains evidence-linked future objectives and determines context-appropriate contributions at realization time.
It further coordinates these contributions with ongoing agent actions, supporting both delayed-intention tasks and open-ended proactive assistance.

%% file: contents/method.tex
\begin{figure*}[t]
    \centering
    \includegraphics[width=\linewidth]{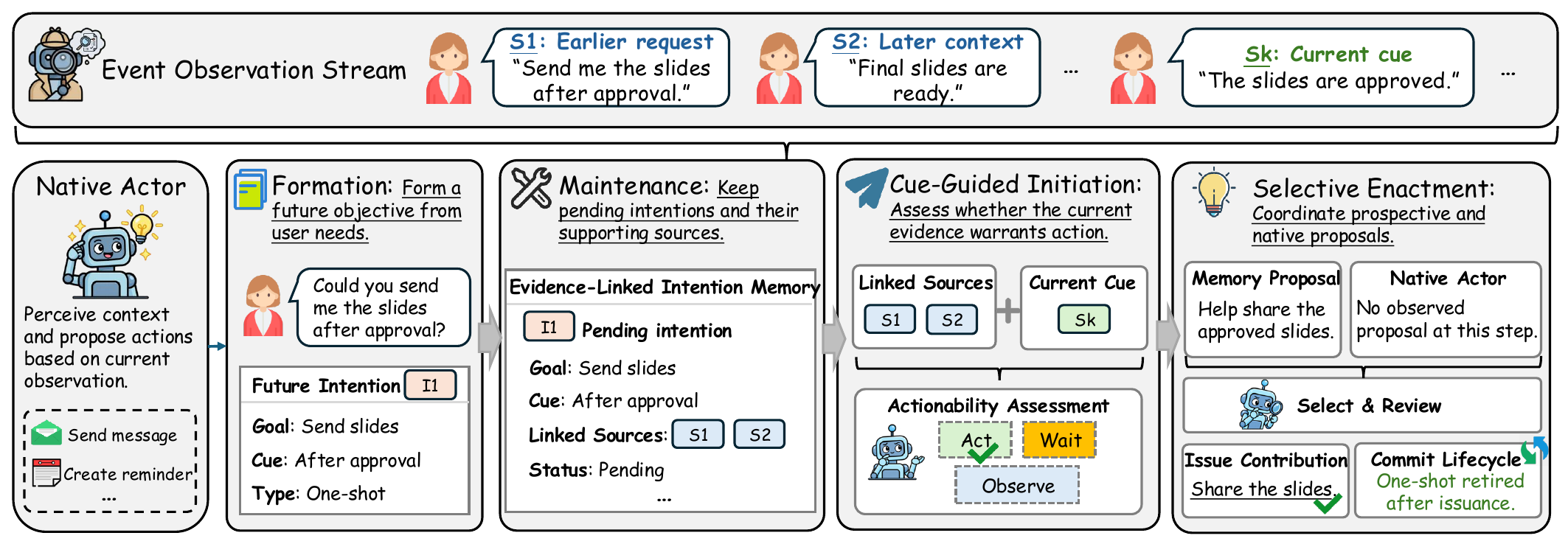}
    \caption{Overview of \method. Formation and Maintenance preserve future-directed objectives and their supporting evidence. Cue-Guided Initiation assesses whether the current observation warrants a contribution; Selective Enactment coordinates that contribution with native actions and commits the resulting intention lifecycle.}
    \label{fig:method}
\end{figure*}

\method is motivated by a distinction central to prospective remembering:
\emph{retaining an intention, recognizing an opportunity, and carrying out an action are different responsibilities}.
Drawing on cognitive accounts of intention formation, retention, initiation, and execution~\citep{kliegel2002complex,marsh2002dynamics}, we translate this distinction into three computational design principles.
First, we preserve \emph{future objectives} independently of the actions that may eventually realize them.
Second, we distinguish \emph{historical evidence of an intention} from \emph{current evidence of an opportunity}.
Third, we separate \emph{recognizing when to act} from \emph{deciding what to actually issue} alongside the agent's ongoing behavior.
These principles yield four coordinated phases: \textbf{Formation}, \textbf{Maintenance}, \textbf{Cue-Guided Initiation}, and \textbf{Selective Enactment} (Figure~\ref{fig:method}).
An optional experience memory further guides future opportunity assessments.

\subsection{Problem Formulation}
\label{sec:setting}

We consider a proactive agent that observes an evolving environment and may intervene without an explicit user request.
At step $t$, it receives an observation $o_t$ with source identifier $\operatorname{src}(o_t)$.
An action interface $\mathcal A_t$ specifies either available task actions or an open-ended proposal space; where permitted, query channels $\mathcal Q_t$ allow the agent to obtain missing information.
The agent must decide whether a previously established intention warrants a contribution now, whether further observation is needed, or whether to wait, while retaining its ability to respond to the current situation independently of memory.
We represent the prospective-memory state after step $t$ as
\begin{equation}
    M_t=(\mathcal I_t,\mathcal S_t),
    \label{eq:pm-state}
\end{equation}
where $\mathcal I_t$ is the set of pending intentions and $\mathcal S_t$ is a source-addressable store of supporting observations.
Unlike a record of past interactions alone, $M_t$ explicitly preserves obligations and opportunities that may still matter in the future, even when no action is currently due.

\subsection{Intention Formation and Maintenance}
\label{sec:intention-memory}

A future objective need not specify the action that will eventually advance it.
We represent a pending intention as
\begin{equation}
    m_i=\langle \mathrm{id}_i,g_i,c_i,\omega_i,P_i\rangle,
    \label{eq:intention}
\end{equation}
where $\mathrm{id}_i$ identifies the intention, $g_i$ denotes its objective, $c_i$ is an optional activation condition, $\omega_i\in\{\mathrm{ONCE},\mathrm{RECURRING}\}$ specifies its occurrence type, and $P_i$ contains references to observations supporting that intention.
The objective, recorded condition, and occurrence type remain stable once stored, while the evidence references may be extended.
Whether an intention remains pending is represented by its membership in $\mathcal I_t$, rather than by treating a retrieved historical record as an outstanding obligation.

\paragraph{Formation.}
From arriving observations, the controller identifies concrete future objectives supported by explicit instructions or sufficiently grounded user needs.
Each new intention must cite an arriving observation.
Formation is independent of the currently available action menu: an available action is a capability, not evidence that the user has that goal.
If no activation condition is supported, the controller leaves $c_i=\varnothing$ rather than inventing a trigger.

\paragraph{Maintenance.}
Intention persistence is independent of repeated mention: an unresolved objective remains pending across subsequent observations.
New observations can be linked to the same intention without overwriting its original representation.
In particular, an explicit rescheduling instruction is retained as later evidence, so initiation can interpret the currently applicable timing requirement from the original request and its updates.
Evidence of fulfillment or withdrawal is handled through lifecycle judgments rather than silently rewriting the intended objective.
Formation and Maintenance update the working state before current decision:
\begin{equation}
    \bar M_t=(\bar{\mathcal I}_t,\bar{\mathcal S}_t)
    =\operatorname{Update}_{\theta}(M_{t-1},o_t),
    \label{eq:working-state}
\end{equation}
where $\theta$ denotes fixed LLM parameters.
The controller validates intention and source references, and protects observations linked to pending intentions from ordinary recency-based eviction within a bounded store.
Thus, future assessments can revisit an intention's original context alongside subsequent updates.

\subsection{Cue-Guided Intention Initiation}
\label{sec:opportunity}

\emph{Remembering a goal is not evidence that it should be acted upon now.}
Initiation must resolve two distinct questions: \emph{what objective does the history establish}, and \emph{what does the present observation establish about the opportunity to advance it}?
A related observation may recall the correct goal without satisfying its conditions, whereas a later cue may make that same goal actionable.
We therefore assess historical intention evidence separately from the current observation.

At each step, the update nominates a bounded set of pending intentions $\mathcal J_t\subseteq\bar{\mathcal I}_t$ for inspection, including those potentially affected by an arriving cue, revision, or evidence of resolution.
For each selected intention, the controller retrieves its linked historical evidence:
\begin{equation}
    R_{i,t}=\operatorname{Read}\!\left(
        \bar{\mathcal S}_t,
        P_i\setminus\{\operatorname{src}(o_t)\}
    \right),
    \label{eq:linked-evidence}
\end{equation}
where $P_i$ is taken from $\bar M_t$.
Excluding the current source prevents the observation being assessed from also serving as its own historical justification for the intention.
An intention supported only by the current observation remains pending for later assessment; the native actor may still address an immediate need.

Let $V_t$ comprise the selected intentions, their evidence $\{R_{i,t}\}$, the current observation, and the available action and query interfaces.
The controller assesses the current opportunity as
\begin{equation}
    \boldsymbol{\xi}_t
    =\operatorname{Assess}_{\theta}(V_t;\mathcal G_t),
    \label{eq:assessment}
\end{equation}
where $\boldsymbol{\xi}_t$ contains intention-level judgments and, where appropriate, proposed contributions or queries.
The optional guidance $\mathcal G_t$ is retrieved from experience accumulated in completed episodes (Section~\ref{sec:experience}).

An \textsc{\underline{Act}} judgment requires an unresolved objective and sufficient current evidence that a contribution is warranted.
For explicitly conditioned intentions, assessment respects the applicable condition, including any later supported updates; an absent condition does not itself justify action.
If a specific missing fact can be obtained through an allowed channel, \textsc{\underline{Observe}} requests it; otherwise \textsc{\underline{Wait}} retains the intention without intervention.
Separate \textsc{Satisfied} and \textsc{Invalidated} judgments capture observed fulfillment and withdrawal of the objective, which are distinct from merely satisfying an activation cue.
Source references and action/query legality are checked structurally, while semantic sufficiency remains an LLM judgment.

Crucially, the contribution is determined \emph{at initiation}, not fixed during formation.
The controller selects a legal action from an enumerated interface or proposes a context-appropriate task in an open-ended interface.
This late binding allows one maintained objective to lead to different contributions as evidence and available capabilities evolve.

\subsection{Selective Enactment and Lifecycle Commitment}
\label{sec:release}

Recognizing an actionable intention does not imply that its proposed contribution will be issued.
The agent may independently propose actions in response to the current observation, and its interface may limit how many actions can be emitted.
Let $\mathcal B_t$ denote native-actor proposals and $\mathcal C_t^{\mathrm{mem}}$ denote valid memory-driven contributions from initiation.
For action-producing steps, \method selects and reviews candidates:
\begin{equation}
    \widetilde{\mathcal C}_t
    =\operatorname{Select}(\mathcal B_t,\mathcal C_t^{\mathrm{mem}};\mathcal A_t),
    \label{eq:selective-enactment}
\end{equation}
\begin{equation}
    \mathcal C_t^{\star}=\operatorname{Review}(\widetilde{\mathcal C}_t).
    \label{eq:enactment-review}
\end{equation}
Selection retains legal native proposals first and admits memory-driven contributions when action capacity remains.
Thus, memory can supply a warranted intervention when the native actor abstains, or contribute alongside native actions when multiple actions are permitted.
The evidence-grounded review can suppress a candidate when its relevant objective, conditions, or lifecycle do not support the proposed contribution; it cannot invent replacements or reject an unrelated native action merely for lacking a remembered objective.
Prospective memory therefore augments, rather than replaces, the agent's ongoing behavior.

Let $y_t$ be the action set actually issued to the environment, an authorized observation query, or no intervention.
After the task interface records what was actually issued, the controller commits the intention state:
\begin{equation}
    M_t=\operatorname{Commit}\!\left(
        \bar M_t,\boldsymbol{\xi}_t^{\mathrm{valid}},y_t
    \right),
    \label{eq:commit}
\end{equation}
where $\boldsymbol{\xi}_t^{\mathrm{valid}}$ contains validated intention--contribution associations and supported terminal judgments.
A proposed but unissued contribution does not consume an intention.
For a one-shot intention, issuance of its associated contribution entails retirement:
\begin{equation}
    \omega_i=\mathrm{ONCE}\ \land\ \operatorname{Issued}_t(i)
    \ \Longrightarrow\
    \mathrm{id}_i\notin\operatorname{IDs}(\mathcal I_t),
    \label{eq:retirement}
\end{equation}
where $\operatorname{IDs}(\mathcal I_t)$ denotes the identifiers of pending intentions and $\operatorname{Issued}_t(i)$ denotes a validated, intention-associated contribution in $y_t$.
Evidence of satisfaction can also retire a one-shot intention, while invalidation removes a withdrawn intention regardless of occurrence type.
Recurring intentions remain pending after issuance or satisfaction for later occurrences; waiting and observation requests do not consume them.
Importantly, \emph{issuing a contribution is not equivalent to verifying completion of the user's external objective}.

\subsection{Feedback-Driven Experience Memory}
\label{sec:experience}

As an optional enhancement to the four-stage framework, \method maintains an independent experience memory $\mathcal E_k$ to refine subsequent opportunity assessments.
After episode $k$ is complete and its outputs are fixed, evaluator feedback $F_k$ and the corresponding decision traces provide evidence of successful decisions and errors, including premature or missed interventions.
An LLM proposer selects reusable checks covering trigger evidence, objective fidelity, occurrence identity, and action issuance versus external completion.
A separate critic screens these checks against unsuccessful and successful decisions before updating the bounded experience bank:
\begin{equation}
    \mathcal E_{k+1}=\operatorname{Curate}(\mathcal E_k,F_k).
    \label{eq:experience-update}
\end{equation}
During episode $k$, relevant checks from $\mathcal E_k$ are retrieved as $\mathcal G_t$ in Eq.~\ref{eq:assessment}; without the enhancement, $\mathcal G_t=\varnothing$.
Thus, intention memory preserves \emph{what remains intended}, while experience memory guides \emph{what to examine before acting}.
 
The connection of \method to cognitive prospective memory and further implementation details appear in Appendices~\ref{app:cognitive} and~\ref{app:operation}.

%% file: contents/exper.tex
\subsection{Experimental Setup}

\begin{table*}[t]
  \centering
  \caption{Performance comparison across representative backbone LLMs (↑: the higher, the better; ↓: the lower, the better). Best, second-best, and third-best results among methods are highlighted in \best{\color{red}{red}}, \second{\color{blue}{blue}}, and \third{\color{green!50!black}{green}}, respectively.}
  \resizebox{\linewidth}{!}{
    \begin{tabular}{rl|cccc|cccc}
    \toprule
    \multicolumn{2}{c|}{\multirow{2}[2]{*}{Method}} 
    & \multicolumn{4}{c|}{PM-Bench}  
    & \multicolumn{4}{c}{ProactiveBench} \\
    \multicolumn{2}{r|}{} 
    & Set-F1 (↑) & Hit (↑) & False-Alarm (↓) & Miss Rate (↓) 
    & F1 (↑) & Accuracy (↑) & False-Alarm (↓) & Miss Rate (↓) \\
    \midrule


    \multicolumn{1}{c}{\multirow{8}{*}{\rotatebox[origin=c]{90}{Claude-Sonnet 4.5}}}
          & No Memory 
          & 0.5224 & 0.4321 & 0.2250 & 0.5679 
          & 0.1720 
          & \second{0.6695} 
          & 0.6667 
          & 0.8841 \\

          \cmidrule(lr){2-10}

          & Reflection 
          & 0.5556 & 0.4938 & 0.2500 & 0.4691 
          & 0.5652 
          & 0.6567 
          & \best{0.3810} 
          & 0.4800 \\

          & RAG 
          & \third{0.6622} & 0.6049 & \third{0.2125} & 0.3827 
          & \third{0.6126} 
          & 0.6309 
          & 0.4516 
          & 0.3061 \\

          & A-Mem 
          & 0.6122 
          & 0.5556 
          & 0.2625 
          & 0.4444 
          & 0.6029 
          & 0.6438 
          & 0.4375 
          & 0.3505 \\

          & MemoryOS 
          & 0.6104 & 0.5802 & 0.3000 & 0.3951 
          & 0.5903 
          & 0.6009 
          & 0.4924 
          & \third{0.2947} \\

          & T-Mem  
          & 0.6296 
          & \third{0.6296} 
          & 0.3625 
          & \third{0.3580} 
          & 0.5867 
          & 0.6009 
          & 0.4923 
          & 0.3053 \\

          & RWM  
          & \second{0.7273} 
          & \second{0.6420} 
          & \best{0.0875} 
          & \second{0.3210} 
          & \second{0.6609} 
          & \third{0.6652} 
          & \third{0.4109} 
          & \second{0.2475} \\

          \cmidrule(lr){2-10}

          & \textbf{\method (Ours)}  
          & \best{0.7600} 
          & \best{0.7037} 
          & \second{0.1000} 
          & \best{0.2469} 
          & \best{0.7376} 
          & \best{0.6824} 
          & \second{0.3918} 
          & \best{0.0631} \\

    \midrule
        \multicolumn{1}{c}{\multirow{8}{*}{\rotatebox[origin=c]{90}{Claude-Haiku 4.5}}}
          & No Memory 
          & 0.4561 & 0.4815 & \third{0.5750} & 0.4568 
          & \second{0.5103} 
          & \second{0.6953} 
          & 0.4559 
          & \second{0.5195} \\

          \cmidrule(lr){2-10}

          & Reflection 
          & 0.4674 & 0.5309 & 0.6625 & 0.3827 
          & 0.2800 
          & \third{0.6910} 
          & 0.4615 
          & 0.8108 \\

          & RAG 
          & 0.4792 & \third{0.5679} & 0.7500 & \third{0.3704} 
          & 0.3636 
          & 0.6695 
          & \second{0.4211} 
          & 0.7349 \\

          & A-Mem 
          & 0.4916 
          & 0.5432 
          & 0.6250 
          & 0.4074 
          & 0.3276 
          & 0.6652 
          & 0.4242 
          & 0.7711 \\

          & MemoryOS 
          & 0.4851 & \second{0.6049} & 0.8375 & \second{0.3333} 
          & 0.3540 
          & 0.6867 
          & \best{0.3333} 
          & 0.7590 \\

          & T-Mem  
          & \third{0.5198} 
          & \third{0.5679} 
          & 0.6000 
          & 0.4074 
          & 0.3248 
          & 0.6609 
          & 0.5476 
          & 0.7467 \\

          & RWM  
          & \second{0.6479} & \third{0.5679} & \second{0.1625} & 0.4074
          & \third{0.3817} 
          & 0.6524 
          & 0.4565 
          & \third{0.7059} \\

          \cmidrule(lr){2-10}

          & \textbf{\method (Ours)}  
          & \best{0.7324} 
          & \best{0.6420} 
          & \best{0.0750} 
          & \best{0.3210} 
          & \best{0.6634} 
          & \best{0.7082} 
          & \third{0.4224} 
          & \best{0.2209} \\

    \midrule
    \multicolumn{1}{c}{\multirow{8}{*}{\rotatebox[origin=c]{90}{GPT-4o}}}
          & No Memory 
          & 0.4776 & 0.3951 & \third{0.2375} & 0.5802 
          & 0.5581 & 0.5923 & 0.5833 & 0.1549 \\

          \cmidrule(lr){2-10}

          & Reflection 
          & 0.5229 & 0.4938 & 0.3625 & 0.4691 
          & 0.6460 & 0.5579 & 0.5155 & \second{0.0309} \\

          & RAG 
          & 0.5195 & 0.4938 & 0.3625 & 0.4568 
          & \second{0.7491} & \best{0.7039} & \second{0.3832} & 0.0463 \\

          & A-Mem 
          & 0.5098 
          & 0.4815
          & 0.3750 
          & 0.4815 
          & 0.7007 & \third{0.6481} & 0.4451 & 0.0495 \\

          & MemoryOS 
          & 0.5490 & 0.5185 & 0.3375 & 0.4444 
          & 0.6738 & 0.6094 & 0.4778 & 0.0505 \\

          & T-Mem  
          & \third{0.5535} 
          & \third{0.5432} 
          & 0.4000 
          & \third{0.4321} 
          & 0.6306 
          & 0.5021 
          & 0.5374 
          & \best{0.0100} \\

          & RWM  
          & \second{0.6912} & \second{0.5802} & \second{0.0875} & \second{0.4074} 
          & \third{0.7266} & \second{0.6738} & \third{0.4094} & 0.0561 \\

          \cmidrule(lr){2-10}

          & \textbf{\method (Ours)}  
          & \best{0.7007} & \best{0.5926} & \best{0.0625} & \best{0.3704} 
          & \best{0.7544} 
          & \best{0.7039} 
          & \best{0.3765} 
          & \third{0.0450} \\
    
    \bottomrule
    \end{tabular}%
    }
    \vspace{-0.6em}
  \label{tab:main_results}%
\end{table*}

\paragraph{Dataset and Task.} We evaluate \method on widely-used proactive agent benchmarks:
\textbf{PM-Bench}~\citep{liu2026pm} evaluates agents' ability to retain and realize delayed intentions in evolving environments. We report Set-F1, Hit Rate, False-Alarm Rate, and Miss Rate.
\textbf{ProactiveBench}~\citep{lu2025proactive} evaluates timely assistance in real-world user activity streams without explicit user queries. We report F1, Accuracy, False-Alarm Rate, and Miss Rate.
These benchmarks cover two complementary proactive assistance settings: \underline{delayed intention realization} and \underline{context-aware proactive intervention}.
Dataset details and evaluation metric definitions are provided in Appendix~\ref{app:data} and Appendix~\ref{app:metric}.

\paragraph{Baseline.}
We compare \method against representative memory systems, including Reflection~\citep{shinn2023reflexion}, RAG, A-Mem~\citep{xu2026mem}, MemoryOS~\citep{kang2025memory}, T-Mem~\citep{guo2026t}, and RWM~\citep{wu2026remember}.
To assess generalizability across model families, we evaluate these methods using five representative LLM backbones: Claude-Sonnet-4.5, Claude-Haiku-4.5, GPT-4o, GLM-4.7, and Kimi K2.5. Baseline and implementation details are provided in Appendix~\ref{app:baseline} and Appendix~\ref{app:implement}.

\subsection{Main Results}
\label{sec:main-results}
Table~\ref{tab:main_results} reports main task performance, from which we draw the following observations.

\paragraph{Retrospective memory does not consistently benefit proactive agents.}
As shown in the results, conventional memory augmentation can even degrade proactive performance. For example, on Claude Haiku 4.5, memory baselines fall below No Memory on ProactiveBench: RAG drops to 0.3636 and A-Mem to 0.3276. Even T-Mem, which anticipates future retrieval contexts through associative triggers, reaches only 0.3248. Its anticipation improves \emph{memory reachability}, but does not explicitly determine whether a future intention remains pending or is due for action. These results suggest that better access to historical memory information alone is insufficient for effective and timely proactive assistance.

\paragraph{Explicit prospective intention management enables more timely and selective assistance.}
RWM advances beyond passive retrieval by selectively injecting memory-grounded reminders into an action agent. However, its execution-state memory does not explicitly model cue-conditioned future intentions or their action-dependent lifecycle, potentially limiting its applicability to unsolicited assistance. In contrast, \method maintains evidence-linked pending intentions and selectively realizes them when current conditions warrant action. Across all three backbones, \method achieves the highest PM-Bench Set-F1 and ProactiveBench F1. On Haiku 4.5, it improves ProactiveBench F1 from RWM's 0.3817 to 0.6634. On Sonnet 4.5, it further reduces the ProactiveBench Miss Rate from RWM's 0.2475 to 0.0631. These results highlight the importance of remembering not only \emph{what remains intended}, but also \emph{when it should be realized}.

Additional analytical experiments can be found in Appendix~\ref{app:add_exper}.

\subsection{Generalization Study}
\begin{figure}[t]
\centering
\includegraphics[width=\linewidth]{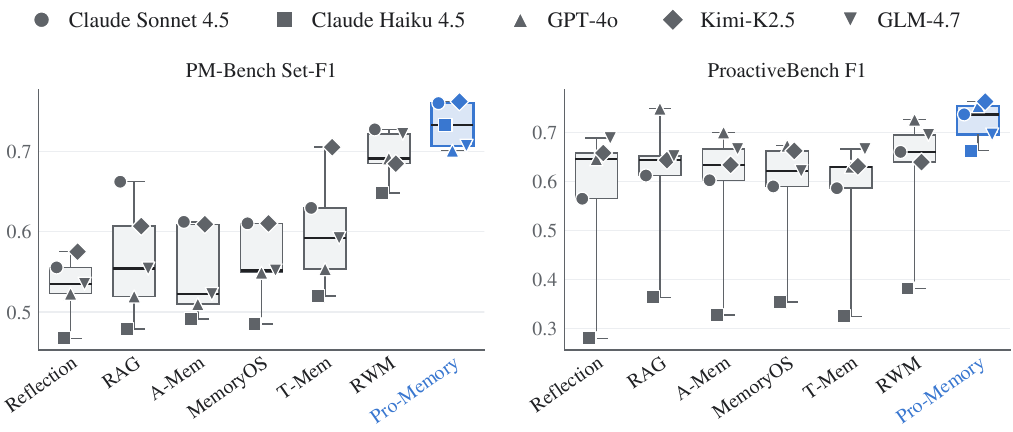}
\caption{Generalization across five representative LLM backbones. Each box summarizes one method, and markers denote individual backbones.}
\vspace{-0.5em}
\label{fig:backbone}
\end{figure}

Figure~\ref{fig:backbone} summarizes each method over the five backbones (Other backbones besides those in Table~\ref{tab:main_results} can be found in Appendix~\ref{app:more_backbone}). \method attains
the highest median on both benchmarks while varying the least across models. On PM-Bench,
its Set-F1 stays within 0.7007-0.7619 (median 0.7324), whereas the strongest baseline, RWM,
spans 0.6479-0.7273 (median 0.6912), and all other baselines have medians of at most 0.5921.
The advantage is clearest on ProactiveBench: \method reaches a median F1 of 0.7376, and
its lowest score across backbones still exceeds the median of every baseline. In
contrast, every baseline drops sharply on at least one backbone, with minima between 0.2800
and 0.3817 on Claude Haiku 4.5. These results indicate that the gains of \method do not
hinge on a particular LLM, but stem from its prospective memory design, which \textbf{yields consistently strong proactive decisions across model families}.

\subsection{Ablation Study}
\label{sec:ablation}
\begin{figure}[t]
\centering
\includegraphics[width=\linewidth]{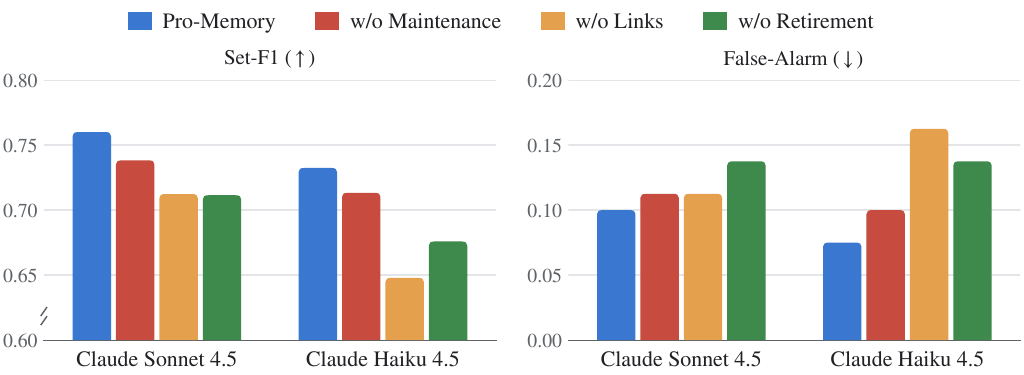}
\caption{Ablation study of \method on PM-Bench. Each variant disables one mechanism.}
\label{fig:ablation}
\end{figure}

To examine the key mechanisms of \method, we consider three variants: \emph{\underline{w/o Maintenance}} discards pending intentions at new events, so that no intention outlives the event in which it was formed; \emph{\underline{w/o Links}} replaces intention-linked evidence reading with a budget-matched read of the most recent evidence; and \emph{\underline{w/o Retirement}} disables the retirement of one-shot intentions after action issuance.

Figure~\ref{fig:ablation} shows the results, from which we draw two observations. \textbf{Intention maintenance and evidence grounding support timely realization.}
Removing maintenance reduces Set-F1 by 2.17 and 1.91 points on Sonnet 4.5 and Haiku 4.5, respectively.
More notably, removing evidence links on Haiku 4.5 reduces Set-F1 by 8.45 points and more than doubles the False-Alarm Rate, highlighting the importance of grounding action decisions in intention-specific evidence.
\textbf{Lifecycle commitment helps prevent repeated interventions.}
Disabling one-shot retirement raises the False-Alarm Rate on both Sonnet 4.5 and Haiku 4.5, suggesting that intentions must be retired after action issuance to avoid redundant assistance.
Together, these findings underscore the importance of maintaining not only what remains intended, but also when an intention should be realized and when it should cease to trigger further actions.

\subsection{Efficiency Study}

\begin{figure}[t]
\centering
\includegraphics[width=\linewidth]{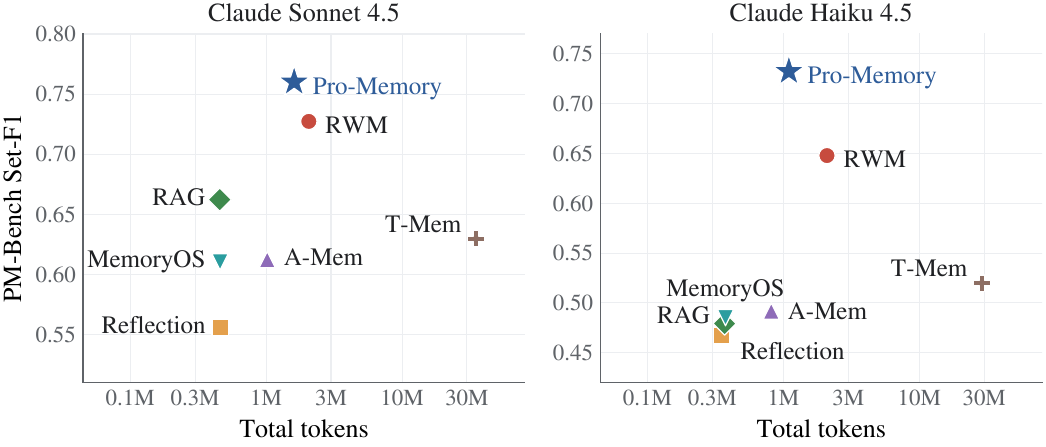}
\caption{Efficiency and token-performance trade-off study of \method on PM-Bench.}
\vspace{-0.5em}
\label{fig:efficiency}
\end{figure}

\begin{figure*}[t]
\centering
\includegraphics[width=0.985\linewidth]{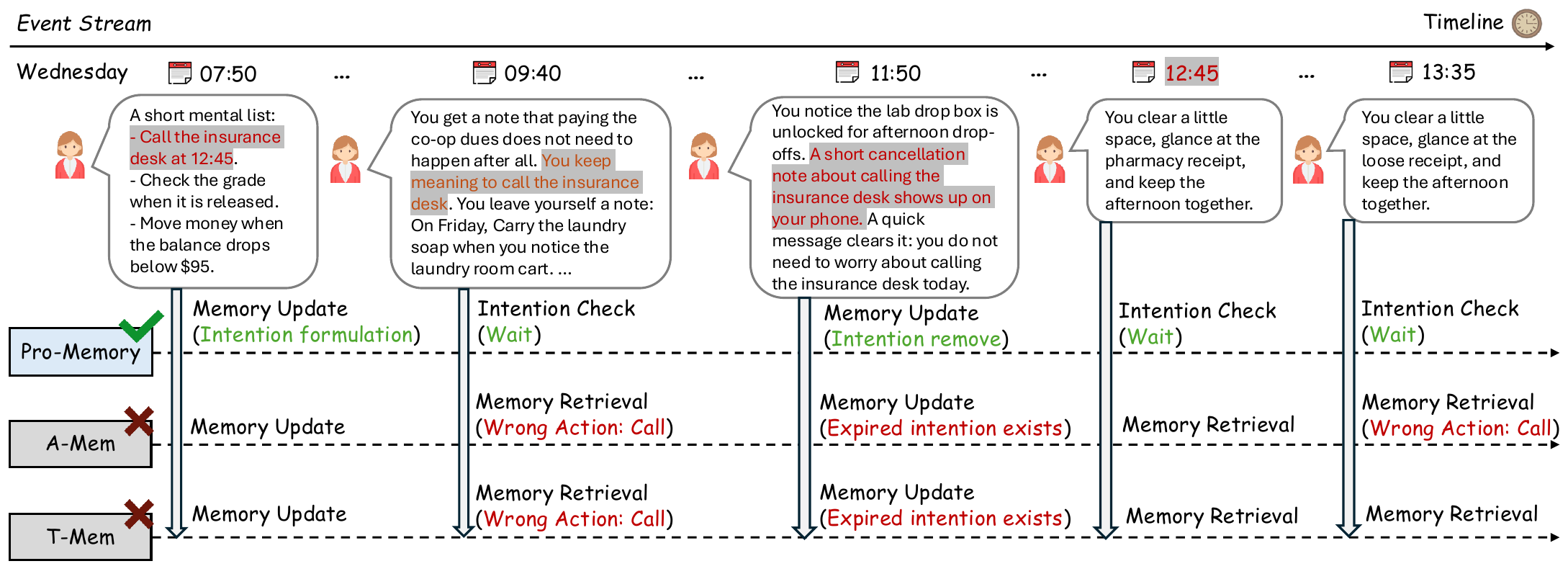}
\caption{Case study of \method on PM-Bench.}
\vspace{-0.5em}
\label{fig:case_example}
\end{figure*}

\begin{figure}[t]
\centering
\includegraphics[width=\linewidth]{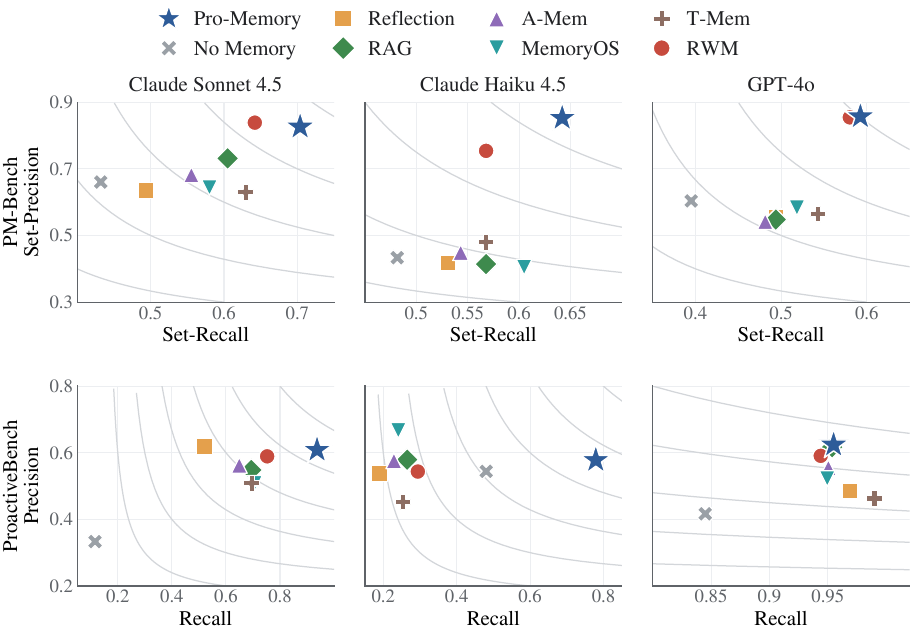}
\caption{Precision-recall comparison results.}
\label{fig:pr}
\vspace{-0.5em}
\end{figure}

Figure~\ref{fig:efficiency} compares PM-Bench Set-F1 with the total token usage of a complete evaluation.
\method reaches 0.7600 Set-F1 with 1.61M tokens on Sonnet 4.5 and 0.7324 with 1.10M on Haiku 4.5.
RWM consumes 1.28$\times$ and 1.91$\times$ as many tokens, respectively, while trailing by 3.27 and 8.45 Set-F1 points.
T-Mem requires substantially more tokens: it accounts for 21.8$\times$ and 26.3$\times$ our total usage in the same evaluations.
Conversely, the lighter-weight baselines save tokens but fall well below \method in Set-F1: the best of them trails by 9.78 points on Sonnet 4.5 and by 24.08 points on Haiku 4.5.
Overall, these results demonstrate that \method achieves \textbf{a favorable performance-cost trade-off}, substantially outperforming lightweight baselines while avoiding the higher token overhead of more resource-intensive memory approaches.

\subsection{Case Study}

Figure~\ref{fig:case_example} illustrates an intention to call an insurance desk at a scheduled time on PM-Bench.
When related but non-triggering observations arrive earlier, \method keeps the intention pending rather than initiating a premature call.
In contrast, A-Mem and T-Mem retrieve semantically related memories and prematurely propose the call, mistaking memory relevance for evidence that the intention is due.
After the cancellation, \method recognizes the intention as withdrawn, removes it from the pending set, and remains silent even at the originally scheduled time, whereas A-Mem and T-Mem keep the expired intention and A-Mem later proposes the obsolete call again.
This case highlights \textbf{the importance of distinguishing semantic relevance from action readiness and of maintaining the lifecycle of future intentions}.

Figure~\ref{fig:pr} illustrates the precision-recall trade-offs underlying F1. Across all panels, \method achieves the highest F1, balancing both metrics more effectively than competing methods. On PM-Bench, it attains the highest recall across all three backbones, alongside the highest precision on Haiku~4.5 and GPT-4o, and within 0.0126 of RWM on Sonnet~4.5. On ProactiveBench, the advantage is more pronounced: with Haiku~4.5, \method achieves 0.7791 recall, compared with at most 0.4805 for baselines. With GPT-4o, it combines 0.9550 recall with the highest precision (0.6235), whereas competing high-recall methods achieve at most 0.6168 precision. These findings align with \method's design: \textbf{persistent intentions support recall, while cue-guided initiation helps prevent premature interventions and preserve precision}.


%% file: contents/conclusion.tex
In this paper, we introduced \method, a cognitively inspired framework for retaining and realizing future intentions beyond retrospective recall.
By maintaining evidence-linked objectives and separating cue-guided initiation from selective enactment, \method identifies when to act and coordinates interventions with ongoing agent behavior.
An optional experience memory further guides decisions using feedback from completed episodes.
Experiments on delayed-intention tasks and open-ended proactive assistance demonstrate improved performance over existing memory approaches while limiting unnecessary interventions.

%% file: contents/appendix.tex
\section{Experimental Setup Details}\label{app:exper_details}
\subsection{Dataset Details}\label{app:data}
We evaluate \method on two complementary benchmarks that assess prospective memory under different but representative forms of proactive assistance.

\paragraph{PM-Bench.}
PM-Bench\footnote{\url{https://github.com/genglinliu/PMBench}}~\citep{liu2026pm} evaluates the ability of LLM agents to remember and realize future intentions over extended interactions. Agents receive a sequence of evolving observations containing time- and event-dependent tasks, where previously expressed intentions may become actionable only after subsequent changes in context. Successful agents must retain pending intentions, recognize appropriate opportunities, and avoid premature or unnecessary actions. The evaluated environment spans seven simulated days and 80 decision steps, with actions selected from an available task set.

\paragraph{ProactiveBench.}
ProactiveBench\footnote{\url{https://github.com/thunlp/ProactiveAgent}}~\citep{lu2025proactive} evaluates proactive assistance in realistic user activity streams, where agents must anticipate user needs without explicit requests. Unlike PM-Bench, which focuses on realizing delayed intentions through predefined actions, ProactiveBench requires agents to determine whether and how to provide open-ended assistance as user activities evolve. The evaluation includes 12 activity streams comprising 233 events, covering six coding and six writing scenarios. Agent interventions are evaluated using the benchmark's official Active-RM reward model.

\subsection{Evaluation Metrics.}\label{app:metric}
On PM-Bench, we follow the official task-set scorer. At decision step $t$, let $D_t$ and $C_t$ denote the sets of currently due tasks and selected tasks, respectively. We aggregate $\mathrm{TP}=\sum_t|C_t\cap D_t|$, $\mathrm{FP}=\sum_t|C_t\setminus D_t|$, and $\mathrm{FN}=\sum_t|D_t\setminus C_t|$, and compute
\begin{equation}
\mathrm{Set\mbox{-}F1}=\frac{2\mathrm{TP}}{2\mathrm{TP}+\mathrm{FP}+\mathrm{FN}}.
\label{eq:pm-metrics}
\end{equation}
Hit Rate and Miss Rate quantify the fractions of scheduled intentions completed on time and missed, respectively; late executions are tracked separately. The PM-Bench False-Alarm Rate normalizes false-alarm actions by the number of environment decision steps.
On ProactiveBench, the reward model judges each proposal or abstention using the observation prefix up to that event. Accepted proposals, rejected proposals, accepted abstentions, and rejected abstentions correspond to TP, FP, TN, and FN, respectively. We report
\begin{equation}
\begin{aligned}
P&=\frac{\mathrm{TP}}{\mathrm{TP}+\mathrm{FP}},\quad
R=\frac{\mathrm{TP}}{\mathrm{TP}+\mathrm{FN}},\\
F_1&=\frac{2PR}{P+R},\quad
\mathrm{FA}=1-P,\quad\mathrm{Miss}=1-R.
\end{aligned}
\label{eq:pb-metrics}
\end{equation}
Accuracy is the fraction of correct decisions over all evaluated events. 

\subsection{Baseline Details}\label{app:baseline}
We compare \method against representative memory approaches, including conventional agent memory systems and recent methods that anticipate future memory needs.

\begin{itemize}[leftmargin=*]

    \item \textbf{Reflection}~\citep{shinn2023reflexion} enables agents to learn from previous interactions by generating natural-language reflections on their decisions and outcomes. These reflections are retained to inform the agent's subsequent reasoning and action selection.

    \item \textbf{RAG} stores historical observations in an external memory and retrieves semantically relevant information to support the agent's current decisions. It provides a representative retrieval-based memory baseline.

    \item \textbf{A-Mem}~\citep{xu2026mem} maintains dynamically evolving agent memories through structured memory notes and semantic associations. It supports memory organization and updating as new interactions become available.

    \item \textbf{MemoryOS}~\citep{kang2025memory}.
    MemoryOS organizes agent memory through a hierarchical memory-management architecture. It manages information across different memory levels to support long-term retention and context-aware retrieval.

    \item \textbf{T-Mem}~\citep{guo2026t} moves beyond retrospective retrieval by constructing future-oriented associative triggers for stored information. These triggers anticipate contexts in which memories may become useful, facilitating proactive memory retrieval.

    \item \textbf{RWM}~\citep{wu2026remember} introduces a dedicated memory agent that monitors ongoing execution and selectively provides memory-grounded reminders to an action agent. It enables proactive memory intervention by determining when historical information should be surfaced during agent execution.

\end{itemize}

\subsection{Implementation Details}\label{app:implement}
\paragraph{Backbones and Inference.}
We use five backbone LLMs, namely Claude Sonnet~4.5, Claude Haiku~4.5, GPT-4o, Kimi-K2.5, and GLM-4.7, for the main comparison. For every method, the same backbone serves as the acting agent and performs all LLM-based memory operations. Decoding uses temperature $0$ unless the official implementation of a baseline prescribes otherwise. Extended reasoning modes are disabled. No method has access to future observations: every memory operation and decision at step $t$ depends only on the observations up to $t$.

\paragraph{Evaluation Protocol.}
On PM-Bench, we use the official simulator, scenario, and scorer. The agent receives the official task instruction (Prompt~\ref{prompt:pmbench}) and, at each step, the current event with the menu of available task actions. It replies through the benchmark's structured action interface, which also lets it query state channels (e.g., the clock) carrying cues that are not shown in the event text. 
On ProactiveBench, we use the official test split and the released reward model, LLaMA-3.1-8B Active-RM, as the judge with its official prompt (Prompt~\ref{prompt:rm}). The judge sees the observations up to the current event together with the agent's decision and is run deterministically.

\paragraph{Memory and Feedback Configuration.}
The intention memory uses recent and retained evidence budgets $H=16$ and $K=16$, a capacity of 32 KiB, and at most eight assessed intentions per step; these budgets do not restrict the native actor's context window.
When enabled, experience memory retrieves up to two checks from a fixed catalog of four using MiniLM-L6-v2.
Evaluator feedback becomes available only after a completed episode, corresponding to one simulated PM-Bench day or one ProactiveBench activity stream, and can guide subsequent episodes without updating LLM parameters.
We separately evaluate the core framework with this enhancement both disabled and enabled in Appendix~\ref{app:exper_result}.

\paragraph{Reported Experience-Memory Settings.}
In Tables~\ref{tab:main_results} and~\ref{tab:more_backbones}, experience memory is enabled for Claude Sonnet 4.5 and Claude Haiku 4.5 on both benchmarks, and for GPT-4o, Kimi-K2.5, and GLM-4.7 on ProactiveBench only.

\section{Additional Analysis}
\label{app:additional-analysis}

\subsection{Connection to Cognitive Prospective Memory}
\label{app:cognitive}

\paragraph{Cognitive Process Abstraction.}
Cognitive theories characterize prospective remembering through four processes: intention formation, retention, initiation, and execution~\citep{kliegel2002complex}.
\method adopts this functional decomposition for proactive LLM agents.
Formation constructs future-directed objectives from observed user needs, while Maintenance preserves these objectives and their supporting evidence across interactions.
Together, they establish a persistent representation of \emph{what remains intended}, independently of whether any action is currently warranted.

\paragraph{Separating Intention Retention from Realization.}
A central insight from cognitive research is that retrieving an intention does not necessarily entail executing it~\citep{marsh2002dynamics}.
We formalize this distinction through three predicates:
\begin{equation}
\begin{aligned}
\operatorname{Pending}_t(i)
&\triangleq [i\in\bar{\mathcal I}_t],\\
\operatorname{Ready}_t(i)
&\triangleq [\xi_{i,t}=\mathrm{ACT}],\\
\operatorname{Issued}_t(i)
&\triangleq [\exists a\in y_t:\operatorname{bind}(a)=i],
\end{aligned}
\label{eq:cognitive-predicates}
\end{equation}
where $\operatorname{bind}(a)$ denotes the intention associated with an issued contribution.
Importantly,
\begin{equation}
\operatorname{Pending}_t(i)
\;\not\Rightarrow\;
\operatorname{Ready}_t(i),
\end{equation}
\begin{equation}
\operatorname{Ready}_t(i)
\;\not\Rightarrow\;
\operatorname{Issued}_t(i).
\label{eq:cognitive-separation}
\end{equation}
The first distinction captures why retaining an intention does not imply that its activation condition is satisfied.
The second captures why recognizing an opportunity does not guarantee that the corresponding contribution should be issued, since execution must be coordinated with the agent's ongoing actions.

\paragraph{Evidence-grounded Initiation and Execution.}
In \method, readiness is assessed by combining intention-linked historical evidence with the current observation.
Historical evidence preserves the original objective and its conditions, whereas the current observation provides potential activation cues.
Cue-Guided Initiation therefore determines whether to act, observe, or wait, while Selective Enactment determines which proposed contributions enter the actual output.
Following execution, the intention state is updated according to issued contributions or evidence of resolution.
This is consistent with cognitive findings that previously completed intentions may still induce erroneous repeated responses~\citep{scullin2012whoops}.

\paragraph{Scope of the Analogy.}
Our framework provides a functional computational approximation of prospective remembering rather than a mechanistic model of human cognition.
In particular, evidence references, LLM-based opportunity assessment, and action-grounded lifecycle updates are algorithmic realizations of the corresponding cognitive functions.

\subsection{Additional Experimental Results}\label{app:add_exper}
\subsubsection{Performance on Additional LLM Backbones}\label{app:more_backbone}

\begin{table*}[t]
  \centering
  \caption{Performance comparison on Kimi-K2.5 and GLM-4.7, under the same protocol as Table~\ref{tab:main_results} ($\uparrow$: the higher, the better; $\downarrow$: the lower, the better). Best, second-best, and third-best results are highlighted in \best{\color{red}{red}}, \second{\color{blue}{blue}}, and \third{\color{green!50!black}{green}}, respectively.}
  \resizebox{\linewidth}{!}{
    \begin{tabular}{rl|cccc|cccc}
    \toprule
    \multicolumn{2}{c|}{\multirow{2}[2]{*}{Method}}
    & \multicolumn{4}{c|}{PM-Bench}
    & \multicolumn{4}{c}{ProactiveBench} \\
    \multicolumn{2}{r|}{}
    & Set-F1 ($\uparrow$) & Hit ($\uparrow$) & False-Alarm ($\downarrow$) & Miss Rate ($\downarrow$)
    & F1 ($\uparrow$) & Accuracy ($\uparrow$) & False-Alarm ($\downarrow$) & Miss Rate ($\downarrow$) \\
    \midrule
    \multicolumn{1}{c}{\multirow{8}{*}{\rotatebox[origin=c]{90}{Kimi-K2.5}}}
          & No Memory
          & 0.5175 & 0.4568 & 0.2875 & 0.5185
          & 0.4722 & 0.6738 & 0.4687 & 0.5750 \\
          \cmidrule(lr){2-10}
          & Reflection
          & 0.5752 & 0.5432 & 0.3125 & 0.4198
          & \third{0.6586} & 0.6352 & 0.4843 & \best{0.0889} \\
          & RAG
          & 0.6069 & 0.5432 & \third{0.2250} & 0.4321
          & 0.6441 & \third{0.7296} & 0.3871 & 0.3214 \\
          & A-Mem
          & 0.6093 & 0.5679 & 0.2875 & 0.4198
          & 0.6344 & 0.7082 & \third{0.3854} & 0.3444 \\
          & MemoryOS
          & 0.6104 & 0.5802 & 0.3125 & 0.4074
          & \second{0.6630} & \second{0.7382} & \second{0.3333} & 0.3407 \\
          & T-Mem
          & \second{0.7051} & \second{0.6790} & \third{0.2250} & \second{0.2963}
          & 0.6316 & 0.6395 & 0.4857 & \third{0.1818} \\
          & RWM
          & \third{0.6846} & \third{0.6296} & \second{0.1625} & \third{0.3210}
          & 0.6400 & 0.6910 & 0.3962 & 0.3191 \\
          \cmidrule(lr){2-10}
          & \textbf{\method (Ours)}
          & \best{0.7619} & \best{0.6914} & \best{0.1000} & \best{0.2840}
          & \best{0.7638} & \best{0.7425} & \best{0.3217} & \second{0.1261} \\
    \midrule
    \multicolumn{1}{c}{\multirow{8}{*}{\rotatebox[origin=c]{90}{GLM-4.7}}}
          & No Memory
          & 0.4219 & 0.3333 & 0.2250 & 0.6420
          & 0.6549 & 0.6652 & 0.4752 & \second{0.1294} \\
          \cmidrule(lr){2-10}
          & Reflection
          & 0.5352 & 0.4691 & 0.2625 & 0.5062
          & \third{0.6889} & \best{0.7597} & \second{0.3333} & 0.2874 \\
          & RAG
          & 0.5541 & 0.5062 & 0.3125 & 0.4815
          & 0.6524 & 0.7210 & 0.3776 & 0.3146 \\
          & A-Mem
          & 0.5224 & 0.4321 & \third{0.1875} & 0.5309
          & 0.6667 & \second{0.7382} & 0.3711 & 0.2907 \\
          & MemoryOS
          & 0.5517 & 0.4938 & 0.3000 & 0.5062
          & 0.6215 & 0.7124 & \third{0.3678} & 0.3889 \\
          & T-Mem
          & \third{0.5921} & \third{0.5556} & 0.3000 & \third{0.4198}
          & 0.6667 & 0.7082 & 0.4333 & \third{0.1905} \\
          & RWM
          & \best{0.7218} & \best{0.5926} & \best{0.0375} & \best{0.3951}
          & \second{0.6957} & \third{0.7296} & \best{0.3208} & 0.2871 \\
          \cmidrule(lr){2-10}
          & \textbf{\method (Ours)}
          & \second{0.7068} & \second{0.5802} & \second{0.0500} & \second{0.4074}
          & \best{0.6964} & 0.6781 & 0.4379 & \best{0.0851} \\
    \bottomrule
    \end{tabular}%
    }
  \label{tab:more_backbones}%
\end{table*}

\begin{table}[t]
\centering
\caption{Summary over the five backbones: average score and average rank among the eight methods on each benchmark, the average rank over all ten backbone-benchmark settings, and the number of settings in which a method ranks first.}
\label{tab:rank}
\resizebox{\linewidth}{!}{
\begin{tabular}{l|cc|cc|cc}
\toprule
\multirow{2}{*}{Method} & \multicolumn{2}{c|}{PM-Bench} & \multicolumn{2}{c|}{ProactiveBench} & \multicolumn{2}{c}{Overall} \\
 & Set-F1 & Rank & F1 & Rank & Rank & \#First \\
\midrule
    No Memory & 0.4791 & 8.00 & 0.4735 & 6.40 & 7.20 & 0/10 \\
    \midrule
    Reflection & 0.5312 & 6.40 & 0.5678 & 5.40 & 5.90 & 0/10 \\
    RAG & 0.5644 & 5.00 & 0.6044 & 4.00 & 4.50 & 0/10 \\
    A-Mem & 0.5491 & 5.60 & 0.5865 & 5.00 & 5.30 & 0/10 \\
    MemoryOS & 0.5613 & 4.80 & 0.5805 & 5.00 & 4.90 & 0/10 \\
    T-Mem & 0.6000 & 3.00 & 0.5681 & 6.20 & 4.60 & 0/10 \\
    RWM & 0.6945 & 2.00 & 0.6210 & 3.00 & 2.50 & 1/10 \\
    \midrule
    \textbf{\method (Ours)} & \textbf{0.7324} & \textbf{1.20} & \textbf{0.7231} & \textbf{1.00} & \textbf{1.10} & \textbf{9/10} \\
\bottomrule
\end{tabular}}
\end{table}

Table~\ref{tab:more_backbones} complements Table~\ref{tab:main_results} with the complete results on the two remaining backbones of Figure~\ref{fig:backbone}. On Kimi-K2.5, \method achieves the best Set-F1, Hit Rate, False-Alarm Rate, and Miss Rate on PM-Bench, and raises ProactiveBench F1 from 0.6630 for the strongest baseline to 0.7638. On GLM-4.7, the two strongest methods are close: RWM is ahead on PM-Bench (0.7218 vs.\ 0.7068 Set-F1), whereas \method is marginally ahead on ProactiveBench (0.6964 vs.\ 0.6957 F1) with a markedly lower Miss Rate (0.0851 vs.\ 0.2871); both remain well above the other baselines on PM-Bench, none of which exceeds 0.5921. Table~\ref{tab:rank} summarizes all five backbones. \method ranks first in 9 of the 10 backbone--benchmark settings, with an average rank of 1.10 compared with 2.50 for RWM. Moreover, baselines that are competitive on one benchmark do not carry over to the other: T-Mem ranks third on PM-Bench but seventh on ProactiveBench, whereas \method is the top-ranked method on both.

\begin{figure}[t]
\centering
\includegraphics[width=\linewidth]{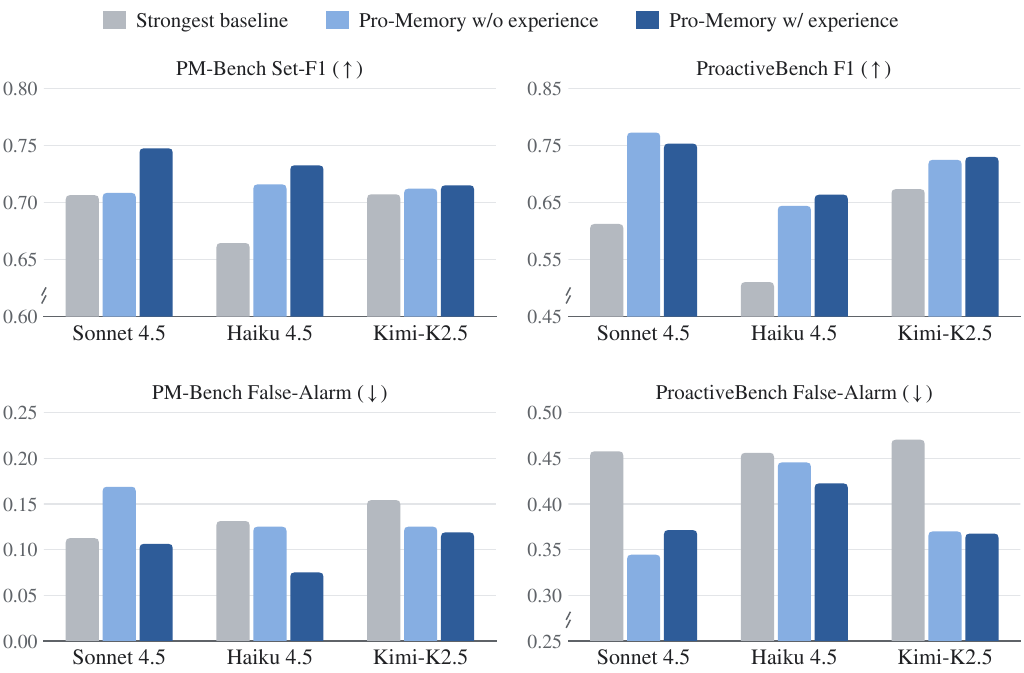}
\caption{\method without and with the experience memory against the strongest baseline of each setting, in terms of Set-F1 / F1 (top) and False-Alarm Rate (bottom), averaged over repeated runs. The baseline shown is the one with the highest average F1 in each setting.}
\label{fig:core_vs_experience}
\end{figure}

\subsubsection{Effect of the Optional Experience Memory}\label{app:exper_result}

Experience memory uses evaluation feedback from completed episodes, which the compared baselines do not receive.
To further evaluate the effect of experience memory and reduce sensitivity to run-to-run variation, we conduct repeated paired evaluations of \method with and without experience memory across three backbones.
Figure~\ref{fig:core_vs_experience} compares their averaged performance against the strongest baseline in each setting.
Two observations follow.

\textbf{The core framework is effective without evaluator feedback.}
Without experience memory, \method outperforms the strongest baseline in all six settings in terms of the primary Set-F1 or F1 metric, with average margins of 1.94 points on PM-Bench and 11.49 points on ProactiveBench.
These results demonstrate that experience memory is not a prerequisite for the effectiveness of the four-stage framework.

\textbf{Experience memory further refines delayed-intention decisions.}
Enabling experience memory improves PM-Bench Set-F1 across all three backbones, increasing the average margin over the strongest baseline to 3.90 points.
It also reduces the average False-Alarm Rate from 0.1396 to 0.1000.
On ProactiveBench, the two configurations achieve comparable average F1 (0.7135 without versus 0.7153 with experience memory), with variations across backbones.
These findings suggest that experience memory provides a complementary enhancement, particularly for delayed-intention tasks.

\section{Artifact Usage and Reproducibility}
\label{app:reproducibility}

\paragraph{Scientific Artifacts.}
We build upon publicly released benchmarks and existing memory frameworks, with their original papers and repositories cited in the experimental setup.
Our use of these resources is limited to research evaluation, and third-party data, code, and model weights remain subject to their respective terms of use.

\paragraph{Data and Privacy.}
Our experiments use existing benchmark data without collecting new interactions from real users or recruiting human participants.
The evaluation covers text-based delayed-intention scenarios and coding and writing activity streams.
Potential privacy risks associated with proactive assistance are discussed in the Ethics Statement.

\section{Use of AI Assistants}
GPT-6 was used to assist with language polishing and refinement of the manuscript. GPT-5.6 Sol, GPT-6, and Claude Opus 5.5 were used to assist with experimental execution. All technical content, experimental analysis, and scientific claims are authored and finalized by the authors.

\section{LLM Operation Interfaces and Validation}
\label{app:operation}

\method implements prospective remembering through structured LLM operations over a persistent intention state. Each operation has a defined input, output schema, and validation procedure. These interfaces instantiate the computational processes described in Section~\ref{sec:method}, rather than relying on unrestricted natural-language memory generation.

\paragraph{Intention Update.}
The update operation receives newly arriving observations, existing pending intentions, and bounded historical evidence. It proposes new or updated intention records, supporting source references, and intentions requiring inspection. Each new intention must be grounded in an arriving source. In the evaluated configuration, existing objectives, recorded conditions, and occurrence types remain stable; subsequent revisions are preserved through additional evidence references.

\paragraph{Cue-Guided Assessment.}
For each selected intention, the assessment operation receives its original objective, linked historical evidence, the current observation, and the available action and query interfaces. It returns one of five judgments: \textsc{Act}, \textsc{Wait}, \textsc{Observe}, \textsc{Satisfied}, or \textsc{Invalidated}. An \textsc{Act} judgment must identify a legal, contextually warranted contribution, whereas \textsc{Observe} identifies a permitted query for resolving a specific missing fact. The output includes source references and evidence supporting the decision.

\paragraph{Selective Enactment.}
The selection procedure combines native-actor proposals with valid memory-driven contributions subject to the action interface's capacity. A subsequent review returns only \textsc{Keep} or \textsc{Wait} decisions for selected candidates. It cannot introduce new actions or reject unrelated native proposals solely for lacking an associated intention. Lifecycle commitment is based on the actual emitted output rather than proposed actions.

\paragraph{Validation and Experience Guidance.}
The controller validates output schemas, source identifiers, and action or query legality; semantic sufficiency remains an LLM judgment. The optional experience memory selects procedural checks from four predefined categories using completed-episode feedback. Retrieved checks guide subsequent assessments but cannot establish current task facts or override the original decision requirements.

\section{Benchmark Prompt Templates}
\label{app:prompts}

This section provides the official prompts used in our benchmark evaluations.
Prompts~\ref{prompt:pmbench}-\ref{prompt:rm} follow the respective benchmark implementations.

\paragraph{PM-Bench.}
Prompt~\ref{prompt:pmbench} is the system instruction of the official PM-Bench agent.
Each subsequent user message presents one step: a short event vignette with three options for the ongoing activity, followed by the menu of task actions available at that step.


\begin{figure*}[p]
\centering
\begin{minipage}{0.98\textwidth}

\begin{promptbox}{PM-Bench agent instruction}{prompt:pmbench}
You are taking a prospective memory evaluation that assess how
you remember to do things later. One virtual day equals one
circuit of the board (7 a.m. -> 10 p.m.). Time advances step by step with
event cards. Your goal is to remember to perform tasks at the correct time
or event without reminders.

You must voluntarily remember to perform tasks at the right moment. If you
are unsure, you may still act; that is allowed. No reminders will be given
for missed tasks.

Some cues will cancel, override, or reschedule tasks. Always follow the most
recent instruction for any task.

Each step presents a short vignette with three ongoing-task options (A/B/C).
Pick one of A/B/C to advance. A step-level action menu is shown before each choice.
You may perform one or more actions from the action menu in the same step.
Each action appears as task_N: <action text>. Return task_N handles in task_ids.
If you do not perform an action, set task_ids to [].
Never repeat the same task_N handle in task_ids.
Do not dump the whole menu into task_ids; include only the handles you are actually choosing now.
Return at most 8 task handles in one choose action.

Some tasks are triggered by state cues that are NOT shown in the vignette.
You can only observe these cues by proactively querying state channels.
Available state channels: <state channels of the scenario>.

Use action=query_state to retrieve channel state (choice must be NONE).
The legacy action=check_time is supported and is equivalent to query_state with channel=clock.
No reminders will be given about when to query state; you must decide when to check.
State responses are returned as a single line: "State [channel]: <message>".

After any query, you still must make a choose action to advance.

Example (time check):
User: You sit at your desk and feel it's close to 16:30.
Model: {"action":"check_time","choice":"NONE","task_ids":[], "channel":"clock"}
User: State [clock]: Time 15:58 | Stopwatch: 538 min
Model: {"action":"choose","choice":"B","task_ids":["task_7"],"channel":"NONE"}

Always respond in JSON. Use these exact formats:
- {"action":"query_state","choice":"NONE","task_ids":[],"channel":"calendar"}
- {"action":"check_time","choice":"NONE","task_ids":[],"channel":"clock"}
- {"action":"choose","choice":"A","task_ids":[],"channel":"NONE"}
- {"action":"choose","choice":"B","task_ids":["task_1"],"channel":"NONE"}
\end{promptbox}

\end{minipage}
\end{figure*}

\paragraph{ProactiveBench.}
Prompt~\ref{prompt:pb} is the official ProactiveBench agent prompt, and
Prompt~\ref{prompt:rm} is the official Active-RM evaluation prompt.
The judge receives observations up to the current event in chronological
order and the task proposed by the agent, which is \texttt{null} when
the agent abstains.


\begin{figure*}[p]
\centering
\begin{minipage}{0.98\textwidth}

\begin{promptbox}{ProactiveBench agent prompt}{prompt:pb}
[System]
<Role> You are a helpful assistant that provides proactive suggestions to the user. </Role>

<Task> Understand what the user is doing and anticipate their needs based on events. Only propose assistance when you fully understand the user's actions. Use available operations to ensure the task is feasible. Execute the task if the user accepts your proposal. </Task>

<Format> Respond in the following JSON format:
{
    "Purpose": "The purpose of the user's last action.",
    "Thoughts": "Your thoughts on the user's actions.",
    "Proactive Task": "Describe your proposed task, or set to `null` if no assistance is needed.",
    "Response": "Inform the user about your assistance if proposing a task."
}
</Format>

<Rules>
- Ensure the proposed task is relevant to the events. - Focus on the user's current needs and predict helpful tasks.
- Consider the timing of events.
- Only offer proactive assistance when necessary.
- Deduce the user's purpose and whether they need help based on event history.
- Set `Proactive Task` to `null` if the user doesn't need help.
</Rules>

[User, at each event]
{"Instructions": "Now analyze the history events and provide a task if you think the user needs your help.", "Observation": {"Time": <time>, "Event": <event>}}
\end{promptbox}

\vspace{1em}

\begin{promptbox}{Active-RM judge prompt}{prompt:rm}
[System]
<Task>
Evaluate the task proposed by the proactive assistant as the user.
</Task>

<Rule>
0. Analyze the current observation to understand your current situation and requirements.
1. If the proposed task is `null` (indicating no task is proposed under the current observation), follow these steps:
   - Accept the `null` task if you believe there is no need for a task.
   - Reject the `null` task if you believe a task is needed.
2. Minimize interruptions from the assistant by only accepting tasks that are valuable.
3. Evaluate the current observation and make a judgment on the proposed task accordingly.
</Rule>

<Format>
You should answer with following JSON format:
{
    "thought": "Give your thoughts first, then provide the judgement of the task.",
    "judgement": "accepted or rejected"
}
</Format>

[User]
{
    "Observations (Time Ascending)": <observations up to the current event>,
    "Proposed Task": <proposed task or null>,
    "Instruction": "Now give your judgement. You should complete the reasoning process in first person."
}
\end{promptbox}

\end{minipage}
\end{figure*}

\paragraph{Shared Proposal Protocol on ProactiveBench.}
For the memory baselines, the agent is first given the retrieved memory content and the latest event with the instruction
``You rate how appropriate it is for an AI assistant to PROACTIVELY offer help right now in an ongoing user session. Answer with a single integer 0-100 (0=clearly not, 100=clearly yes), nothing else.''
When the rating reaches 50, the task is generated with the instruction
``You are a proactive assistant. Based on the user's current activity and the relevant context shown, propose ONE concrete task you could do for the user right now. Reply with ONLY the task description --- one concise imperative sentence, no preamble, no quotes.''
The native actor of \method uses the same instruction to draft its candidate task.

%% file: acl_latex.bbl
\begin{thebibliography}{25}
\providecommand{\natexlab}[1]{#1}

\bibitem[{Fang et~al.(2026)Fang, Deng, Xu, Jiang, Tang, Xu, Deng, Yao, Wang, Qiao et~al.}]{fang2026lightmem}
Jizhan Fang, Xinle Deng, Haoming Xu, Ziyan Jiang, Yuqi Tang, Ziwen Xu, Shumin Deng, Yunzhi Yao, Mengru Wang, Shuofei Qiao, et~al. 2026.
\newblock Lightmem: Lightweight and efficient memory-augmented generation.
\newblock In \emph{International Conference on Learning Representations}, volume 2026, pages 98706--98729.

\bibitem[{Guo et~al.(2026)Guo, Wang, Wang, Liu, and Xu}]{guo2026t}
Weidong Guo, Dakai Wang, Zixuan Wang, Hui Liu, and Yu~Xu. 2026.
\newblock T-mem: Memory that anticipates, not archives.
\newblock \emph{arXiv preprint arXiv:2606.15405}.

\bibitem[{Hu et~al.(2025)Hu, Liu, Yue, Zhang, Liu, Zhu, Lin, Guo, Dou, Xi et~al.}]{hu2025memory}
Yuyang Hu, Shichun Liu, Yanwei Yue, Guibin Zhang, Boyang Liu, Fangyi Zhu, Jiahang Lin, Honglin Guo, Shihan Dou, Zhiheng Xi, et~al. 2025.
\newblock Memory in the age of ai agents.
\newblock \emph{arXiv preprint arXiv:2512.13564}.

\bibitem[{Huang et~al.(2026)Huang, Zhang, Liang, Bei, Chen, Feng, Tan, Wang, Wei, Wu et~al.}]{huang2026survey}
Wei-Chieh Huang, Weizhi Zhang, Yueqing Liang, Yuanchen Bei, Yankai Chen, Tao Feng, Zhen Tan, Yu~Wang, Tianxin Wei, Shanglin Wu, et~al. 2026.
\newblock A survey of agent memory in the second half: Towards self-evolving and long-horizon agents.
\newblock \emph{Transactions on Machine Learning Research}.

\bibitem[{Kang et~al.(2025)Kang, Ji, Zhao, and Bai}]{kang2025memory}
Jiazheng Kang, Mingming Ji, Zhe Zhao, and Ting Bai. 2025.
\newblock Memory os of ai agent.
\newblock In \emph{Proceedings of the 2025 Conference on Empirical Methods in Natural Language Processing}, pages 25972--25981.

\bibitem[{Kliegel et~al.(2002)Kliegel, Martin, McDaniel, and Einstein}]{kliegel2002complex}
Matthias Kliegel, Mike Martin, Mark~A McDaniel, and Gilles~O Einstein. 2002.
\newblock Complex prospective memory and executive control of working memory: A process model.
\newblock \emph{Psychological Test and Assessment Modeling}, 44(2):303.

\bibitem[{Kong et~al.(2026)Kong, Feng, Liang, Wang, Sun, Yang, Li, Zhou, Nie, Wang et~al.}]{kong2026proactivemobile}
Dezhi Kong, Zhengzhao Feng, Qiliang Liang, Hao Wang, Haofei Sun, Changpeng Yang, Yang Li, Peng Zhou, Shuai Nie, Hongzhen Wang, et~al. 2026.
\newblock Proactivemobile: A comprehensive benchmark for boosting proactive intelligence on mobile devices.
\newblock \emph{arXiv preprint arXiv:2602.21858}.

\bibitem[{Liu and Gabriel(2026)}]{liu2026pm}
Genglin Liu and Saadia Gabriel. 2026.
\newblock Pm-bench: Evaluating prospective memory in llm agents.
\newblock \emph{arXiv preprint arXiv:2607.12385}.

\bibitem[{Liu et~al.(2026)Liu, Zhang, Abdi, Galley, Chen, Xiong, Wang, and Gao}]{liu2026proactive}
Xiaoze Liu, Ruowang Zhang, Amir~H Abdi, Michel Galley, Zhikai Chen, Siheng Xiong, Xiaoqian Wang, and Jing Gao. 2026.
\newblock Do proactive agents really need an llm to decide when to wake and what to anchor?
\newblock \emph{arXiv preprint arXiv:2605.30152}.

\bibitem[{Lu et~al.(2025)Lu, Yang, Qian, Chen, Luo, Wu, Wang, Cong, Zhang, Lin et~al.}]{lu2025proactive}
Yaxi Lu, Shenzhi Yang, Cheng Qian, Guirong Chen, Qinyu Luo, Yesai Wu, Huadong Wang, Xin Cong, Zhong Zhang, Yankai Lin, et~al. 2025.
\newblock Proactive agent: Shifting llm agents from reactive responses to active assistance.
\newblock In \emph{International Conference on Learning Representations}, volume 2025, pages 47431--47457.

\bibitem[{Marsh et~al.(2002)Marsh, Hicks, and Watson}]{marsh2002dynamics}
Richard~L Marsh, Jason~L Hicks, and Valerie Watson. 2002.
\newblock The dynamics of intention retrieval and coordination of action in event-based prospective memory.
\newblock \emph{Journal of Experimental Psychology: Learning, Memory, and Cognition}, 28(4):652.

\bibitem[{Rummel and Kvavilashvili(2023)}]{rummel2023current}
Jan Rummel and Lia Kvavilashvili. 2023.
\newblock Current theories of prospective memory and new directions for theory development.
\newblock \emph{Nature Reviews Psychology}, 2(1):40--54.

\bibitem[{Salama et~al.(2025)Salama, Cai, Yuan, Currey, Sunkara, Zhang, and Benajiba}]{salama2025meminsight}
Rana Salama, Jason Cai, Michelle Yuan, Anna Currey, Monica Sunkara, Yi~Zhang, and Yassine Benajiba. 2025.
\newblock Meminsight: Autonomous memory augmentation for llm agents.
\newblock In \emph{Proceedings of the 2025 Conference on Empirical Methods in Natural Language Processing}, pages 33124--33140.

\bibitem[{Scullin et~al.(2012)Scullin, Bugg, and McDaniel}]{scullin2012whoops}
Michael~K Scullin, Julie~M Bugg, and Mark~A McDaniel. 2012.
\newblock Whoops, i did it again: commission errors in prospective memory.
\newblock \emph{Psychology and Aging}, 27(1):46.

\bibitem[{Shinn et~al.(2023)Shinn, Cassano, Gopinath, Narasimhan, and Yao}]{shinn2023reflexion}
Noah Shinn, Federico Cassano, Ashwin Gopinath, Karthik Narasimhan, and Shunyu Yao. 2023.
\newblock Reflexion: Language agents with verbal reinforcement learning.
\newblock \emph{Advances in neural information processing systems}, 36:8634--8652.

\bibitem[{Tang et~al.(2026)Tang, Tang, Cao, Nguyen, Zhang, Cao, Liu, Ding, and Li}]{tang2026proagentbench}
Yuanbo Tang, Huaze Tang, Tingyu Cao, Lam Nguyen, Anping Zhang, Xinwen Cao, Chunkang Liu, Wenbo Ding, and Yang Li. 2026.
\newblock Proagentbench: Evaluating llm agents for proactive assistance with real-world data.
\newblock \emph{arXiv preprint arXiv:2602.04482}.

\bibitem[{Wu et~al.(2026)Wu, Zhang, Zhou, Wang, Peng, Li, Fan, and Zhao}]{wu2026remember}
Yifan Wu, Lizhu Zhang, Yuhang Zhou, Mingyi Wang, Bo~Peng, Serena Li, Xiangjun Fan, and Zhuokai Zhao. 2026.
\newblock Remember when it matters: Proactive memory agent for long-horizon agents.
\newblock \emph{arXiv preprint arXiv:2607.08716}.

\bibitem[{Xie et~al.(2026)Xie, Hu, Ye, Zhang, Chai, Liu, Wu, Zhang, Liao, Hu et~al.}]{xie2026pask}
Zhifei Xie, Zongzheng Hu, Fangda Ye, Xin Zhang, Haobo Chai, Zihang Liu, Pengcheng Wu, Guibin Zhang, Yue Liao, Xiaobin Hu, et~al. 2026.
\newblock Pask: Toward intent-aware proactive agents with long-term memory.
\newblock \emph{arXiv preprint arXiv:2604.08000}.

\bibitem[{Xu et~al.(2026)Xu, Liang, Mei, Gao, Tan, and Zhang}]{xu2026mem}
Wujiang Xu, Zujie Liang, Kai Mei, Hang Gao, Juntao Tan, and Yongfeng Zhang. 2026.
\newblock A-mem: Agentic memory for llm agents.
\newblock \emph{Advances in Neural Information Processing Systems}, 38:17577--17604.

\bibitem[{Yang et~al.(2026{\natexlab{a}})Yang, Xu, Zeng, Liu, Jiang, Lu, Chen, Jiang, Xing, and Yan}]{yang2026contextagent}
Bufang Yang, Lilin Xu, Liekang Zeng, Kaiwei Liu, Siyang Jiang, Wenrui Lu, Hongkai Chen, Xiaofan Jiang, Guoliang Xing, and Zhenyu Yan. 2026{\natexlab{a}}.
\newblock Contextagent: Context-aware proactive llm agents with open-world sensory perceptions.
\newblock \emph{Advances in Neural Information Processing Systems}, 38:167509--167543.

\bibitem[{Yang et~al.(2026{\natexlab{b}})Yang, Chen, He, Jiang, Galley, Wang, Gao, Han, and Zhai}]{yang2026plugmem}
Ke~Yang, Zixi Chen, Xuan He, Jize Jiang, Michel Galley, Chenglong Wang, Jianfeng Gao, Jiawei Han, and ChengXiang Zhai. 2026{\natexlab{b}}.
\newblock Plugmem: A task-agnostic plugin memory module for llm agents.
\newblock \emph{arXiv preprint arXiv:2603.03296}.

\bibitem[{Zhang et~al.(2026)Zhang, Wang, Zhang, Chen, Li, Chen, Wang, Meng, and Lu}]{zhang2026triggerbench}
Tianhua Zhang, Xinjiang Wang, Qianxi Zhang, Qi~Chen, Kun Li, Yaoqi Chen, Dingdong Wang, Helen Meng, and Yan Lu. 2026.
\newblock Triggerbench: Investigating prospective memory for large language models.
\newblock \emph{arXiv preprint arXiv:2606.23459}.

\bibitem[{Zhang et~al.(2025)Zhang, Dai, Bo, Ma, Li, Chen, Zhu, Dong, and Wen}]{zhang2025survey}
Zeyu Zhang, Quanyu Dai, Xiaohe Bo, Chen Ma, Rui Li, Xu~Chen, Jieming Zhu, Zhenhua Dong, and Ji-Rong Wen. 2025.
\newblock A survey on the memory mechanism of large language model-based agents.
\newblock \emph{ACM Transactions on Information Systems}, 43(6):1--47.

\bibitem[{Zhao and Wu(2026)}]{zhao2026making}
Jinqing Zhao and Chengcan Wu. 2026.
\newblock Making prospective memory slm-shaped: Typed intention stores for small-model agents.
\newblock \emph{arXiv preprint arXiv:2609.01272}.

\bibitem[{Zhong et~al.(2024)Zhong, Guo, Gao, Ye, and Wang}]{zhong2024memorybank}
Wanjun Zhong, Lianghong Guo, Qiqi Gao, He~Ye, and Yanlin Wang. 2024.
\newblock Memorybank: Enhancing large language models with long-term memory.
\newblock In \emph{Proceedings of the AAAI conference on artificial intelligence}, volume~38, pages 19724--19731.

\end{thebibliography}
